\documentclass[10pt,twocolumn,twoside]{IEEEtran} 
\usepackage{etex}
\usepackage{siunitx}
\usepackage[compress,nospace]{cite}
\usepackage{subfigure}
\usepackage{amssymb}
\usepackage{algorithm}
\usepackage[noend]{algpseudocode}

\ifCLASSINFOpdf
   \usepackage[pdftex]{graphicx}
\else
\fi

\usepackage[caption=false,font=normalsize,labelfont=sf,textfont=sf]{subfig}
\usepackage{array}
\usepackage{url}
\usepackage[usenames,dvipsnames]{color}
\usepackage{tikz}
\usetikzlibrary{shapes,arrows,shadows,fit}
\usepackage{amssymb,amsthm}
\usepackage[cmex10]{amsmath}
\usepackage{pgfplots}
\pgfplotsset{compat=1.7}

\usepackage{multirow}
\usepackage{xcolor}
\usepackage{colortbl}
\usepackage{hhline}
\usepackage{pbox}

\usepackage{cellspace}
\usepackage{mdframed}
\usepackage{algorithm}
\usepackage{color}
\usepackage[noend]{algpseudocode}

\pgfdeclarelayer{background}
\pgfdeclarelayer{foreground}
\pgfsetlayers{background,main,foreground}
\tikzstyle{sensor}=[draw, fill=blue!20, text width=5em, 
    text centered, minimum height=2.5em,drop shadow]
\tikzstyle{ann} = [above, text width=5em, text centered]
\tikzstyle{wa} = [sensor, text width=10em, fill=red!20, 
    minimum height=6em, rounded corners, drop shadow]
\tikzstyle{sc} = [sensor, text width=13em, fill=red!20, 
    minimum height=10em, rounded corners, drop shadow]

\definecolor{Baseline}{rgb}{1,1,0.88}
\definecolor{P1}{rgb}{0.88,0.88,1}
\definecolor{TwoPriors}{rgb}{0.88,1,0.88}
\definecolor{ThreePriors}{rgb}{1.0,0.88,0.88}

\begin{document}

\title{Multiview Based 3D Scene Understanding On Partial Point Sets}

\author{Ye Zhu*,
        Sven Ewan Shepstone,~\IEEEmembership{Member,~IEEE,}
        Pablo Mart\'inez-Nuevo,~\IEEEmembership{Member,~IEEE,}
        Miklas Str\o m Kristoffersen,~\IEEEmembership{Student Member,~IEEE,}
        Fabien Moutarde,~\IEEEmembership{Member,~IEEE,}
        Zhuang Fu,~\IEEEmembership{Member,~IEEE}
\thanks{Y. Zhu and Z. Fu are with Shanghai Jiao Tong University, 200240 Shanghai, China  (e-mail: szyezhu@gmail.com, zhfu@sjtu.edu.cn).}
\thanks{S.E. Shepstone, P. Mart\'inez-Nuevo and M. Kristoffersen are with Bang and Olufsen A/S, Peter Bangs Vej 15, 7600 Struer, Denmark (e-mail: {ssh, pmn, mko}@bang-olufsen.dk). }
\thanks{F. Moutarde is with Mines ParisTech, 75272 Paris, France (e-mail: fabien.moutarde@mines-paristech.fr).}
\thanks{This work is supported by Bang and Olufsen A/S, Denmark.}
}

%

\maketitle

\begin{abstract}
Deep learning within the context of point clouds has gained much research interest in recent years mostly due to the promising results that have been achieved on a number of challenging benchmarks, such as 3D shape recognition and scene semantic segmentation. In many realistic settings however, snapshots of the environment are often taken from a single view, which only contains a partial set of the scene due to the field of view restriction of commodity cameras. 3D scene semantic understanding on partial point clouds is considered as a challenging task. In this work, we propose a processing approach for 3D point cloud data based on a multiview representation of the existing 360$^\circ$ point clouds. By fusing the original 360$^\circ$ point clouds and their corresponding 3D multiview representations as input data, a neural network is able to recognize partial point sets while improving the general performance on complete point sets, resulting in an overall increase of 31.9\% and 4.3\% in segmentation accuracy for partial and complete scene semantic understanding, respectively. This method can also be applied in a wider 3D recognition context such as 3D part segmentation. 
\end{abstract}

\begin{IEEEkeywords}
partial point cloud, multiview representation, 3D scene understanding, deep learning
\end{IEEEkeywords}

%
\IEEEpeerreviewmaketitle

\section{Introduction}
\label{sec:intro}

Deep learning on point clouds is a rather recent research topic. Regarded as simple, yet unstructured geometric data, point cloud has not attracted much attention till recently when the PointNet series\cite{qi2017pointnet,qi2017pointnet++,qi2017frustum} came out. As the first type of deep neural networks(DNNs) that take a direct point cloud as input, the PointNet series have been a great success in opening a new chapter in this field with wide applications to multiple 3D recognition tasks. Following the work of PointNet, \cite{klokov2017escape} introduces a Kd-network to realize 3D shape recognition on point clouds. PointSIFT is proposed by Jiang et al. \cite{jiang2018pointsift} to further improve a network's performance. These existing works model complete point clouds using several 3D datasets like ModelNet40\cite{wu20153d}, ShapeNet\cite{chang2015shapenet} and Scannet\cite{dai2017scannet} and have rarely looked into the problem of recognizing partial point sets. However, when it comes to real-world application of scene understanding, it is more difficult to construct such a 360$^\circ$ point set of an entire room. Instead, it is more likely to obtain a partial set of points from a snapshot of the indoor environment. Furthermore, a complete scene point set is not always needed since we are often interested in analyzing the local geometric relationships between certain categories of objects instead of general understanding of the entire room.

Compared to other related fields, studies on partial point clouds are limited. Existing works on partial point clouds are primarily concerned with reconstruction of complete point sets based on incomplete ones\cite{mei20173d, previtali2018towards, figueiredo2017automatic}.

\begin{figure}[htb]
\begin{center}
\includegraphics[width=0.95\linewidth]{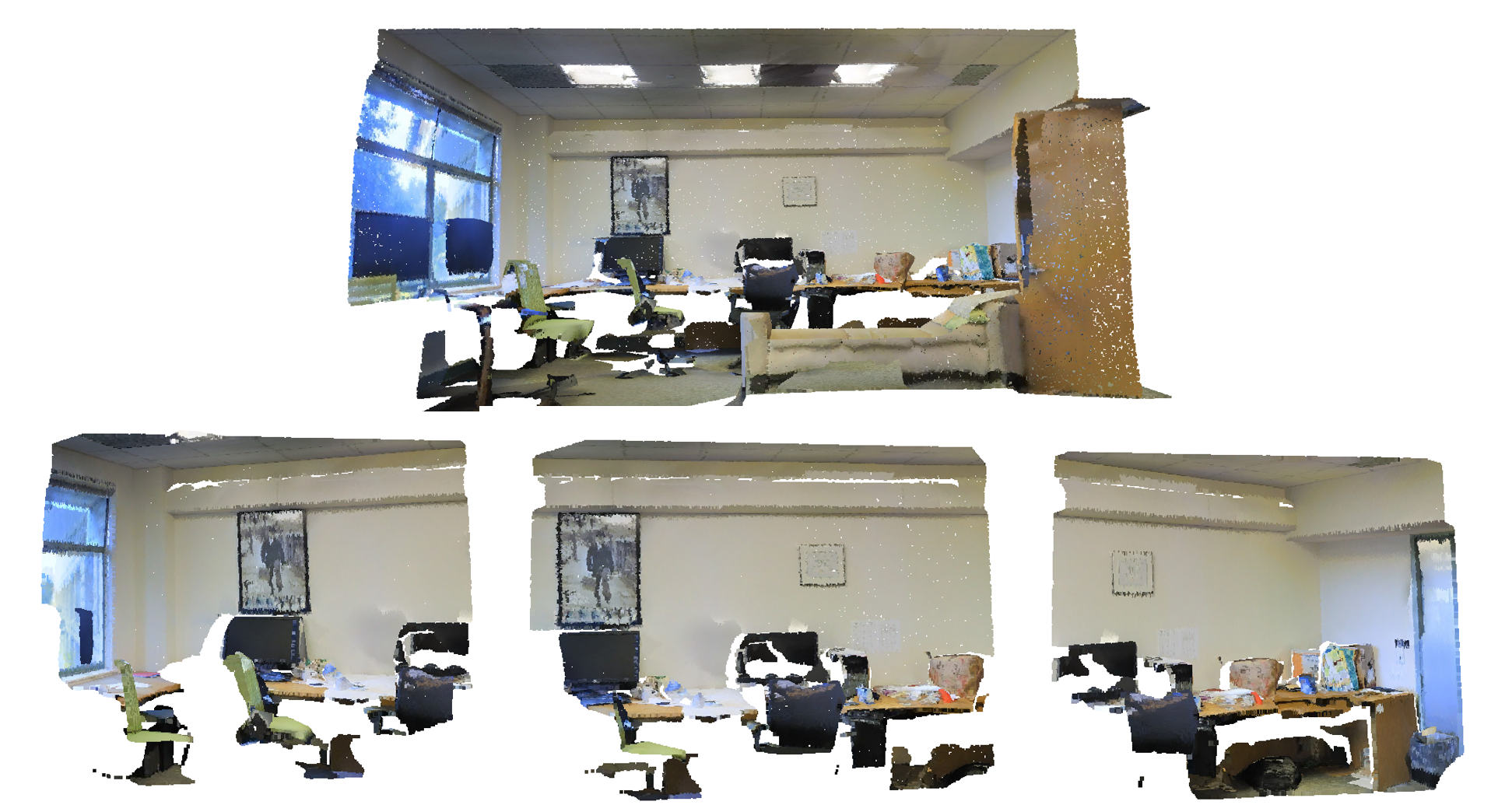}
\end{center}
    \caption{An example of a 3D multiview representation for a complete scene point cloud.
    The point cloud above is an original one from Stanford dataset\cite{armeni_cvpr16}.
    The three point sets at the bottom are chosen from the collection of multiview-based partial point sets. All viewed from the same viewpoint.}
    \label{fig:downsampled_example}
\end{figure}

In this paper, we focus on the problem of 3D scene understanding on partial point sets and propose a point cloud data processing method that seeks to represent a 360$^\circ$ point cloud as multiple multiview-based partial point sets. In other words, this multiview representation is actually a collection of partial point sets extracted from the complete ones. Earlier works on 3D modelling usually tackle the problem through two representation types, which are volumetric\cite{wu20153d, qi2016volumetric,sharma2016vconv} and multiview representation\cite{su2015multi,choy20163d,simon2017hand,chen2017multi,tulsiani2017multi}. Compared to volumetric representation, multiview representation has achieved more competitive performance on 3D recognition tasks. Classic multiview CNNs render 3D models as multiple 2D images and use 2D convolutional networks to accomplish specific tasks. Our work follows in the same vein from Su et al.\cite{su2015multi}, but with the major difference in that, instead of using 2D images from different views for the 3D object, we maintain the 3D property of the original point clouds. 

Although we concentrate mainly on the task of scene understanding, our multiview representation describes a simple and unified approach that is applicable for deep learning on point clouds in general. The core idea is to perform downsampling on the original 360$^\circ$ point cloud from various perspectives. We wish to simulate the partial point sets as the snapshot we would get in a real-world context, which should contain a certain field of view(FOV) restriction of real cameras, with the visibility of points from that FOV.

We provide a concrete approach that realizes this multiview presentation operation and demonstrates its effectiveness through both theoretical and experimental analyses. By supplementing the original training samples with the multiview-based partial point sets as input, we prove that a neural network acquires the ability to recognize partial scene point sets and improves the performance on 360$^\circ$ scene point clouds simultaneously. Further experiments of 3D part segmentation illustrate that our work can be applied in a wider context than scene understanding.

The key contributions of our work are as follows:

\begin{itemize}
\item We propose a preprocessing approach that represents the existing complete 360$^\circ$ point set as a collection of 3D multiview-based partial point sets, and then combine both types of point clouds as input for training;

\item We illustrate both theoretically and experimentally that the deep neural network is able to enhance the ability to recognize both 360$^\circ$ and incomplete partial sets of point clouds through this approach.

\end{itemize}

We avoid referring our approach as an ``data augmentation" method because the emphasis of the work lies on the contribution of 3D multiview representation. Providing the partial data as ``augmentation" is only an easiest way to guarantee the effectiveness of this representation. In addition, we prove in the later experiments that even without the increase on the size of training data, the multiview representation also works.

\section{Problem Formulation}
\label{sec:method}

\subsection{Background}

The main idea of our method is to represent the 360$^\circ$ point cloud as a collection of multiview-based partial sets. 

Intuitively, it is reasonable that a DNN has difficulties in recognizing partial point sets when trained only with 360$^\circ$ complete point clouds, due to the reason that it recognizes a 3D shape $S^{'}$ by a set of critical points $\mathcal{C}^{'}_{S}$ \cite{qi2017pointnet}. Therefore, the incompleteness of partial point sets caused by occlusion leads to certain missing critical points, which makes recognizing the actual category challenging.

Suppose we have a point cloud $S^{'}$ represented as a set of 3D points $\{P_{i}| i = 1,...,n\}$ that the DNN wish to recognize, where $P_{i} \in \mathbb{R}^{3}$. This complete point cloud $S^{'}$ that has been studied in previous works is actually a uniformly downsampled subset of the ideal original shape $S$ which should contain an infinite number of points. By fusing the 360$^\circ$ point data and their 3D multiview representation, we want to achieve two objectives:
On the one hand, we want these additional multiview-based partial point sets $\{S^{''}_{i}\}_{(i = 1,...,w)}$ to help the network to acquire the ability to recognize similar incomplete point sets. On the other hand, we do not want the original performance on complete point sets to be impaired due to this multiview processing operation. Furthermore, it is desirable that the original performance can be further improved.

We choose to demonstrate our idea based on the work of PointNet\cite{qi2017pointnet}. The reason for which we choose PointNet is that it is the first and the most classic DNN for point clouds. Its clear and unified architecture helps to provide a better theoretical understanding for our method. PointNet++ maintains the core architecture of PointNet but incorporates a hierarchical structure that takes the local metric space into consideration\cite{qi2017pointnet++}, which could risk biasing influence on our proposed method.

For a DNN for point clouds like PointNet, the general objective of the network is to approximate  a general function defined on a point set by applying a symmetric function on transformed elements in the set:

\begin{equation}
\label{eq1}
    f(\{x_{1},...,x_{n}\}) \approx g(h(x_{1}),...,h(x_{n}))
\end{equation}

where $f : 2^{\mathbb{R}^{N}} \rightarrow \mathbb{R}$, $h : \mathbb{R}^{N} \rightarrow \mathbb{R}^{K}$ and $g : \underset{n}{\underbrace{\mathbb{R}^{K} \times ... \times \mathbb{R}^{K}}} \rightarrow \mathbb{R}$ is a symmetric function, $x_{i}$ as the points of (\ref{eq1}).

\vspace{6pt}

The following main result from \cite{qi2017pointnet} is important for our development. 

Define $\textbf{u} = \underset{x_{i}\in S}{MAX}\{h(x_{i})\}$  to be the sub-network of $f$ which maps a point set in $\left [ 0,1 \right ]^{m}$ to a K-dimensional vector. $MAX$ is a vector max operator that takes $n$ vectors as input and returns a new vector of the element-wise maximum, and $h$ is a soft indicator function which can be interpreted as the spatial encoding of a point. Let $\chi = \{S:S\subseteq [0,1] $ and $\left | S \right |\ = n\}$, $\gamma$ be a continuous function. The choice of the $MAX$ function is in accordance to the max pooling layer in the architecture of PointNet. K is the bottleneck dimension of $f$, which corresponds to the number of dimension in (\ref{eq1}), it is also the number of neurons in the max pooling layer.

\vspace{6pt}

\textbf{Theorem 1} Suppose $\textbf{u} : \chi \rightarrow \mathbb{R}^{K} such \: that \: \textbf{u} = \underset{x_{i}\in S}{MAX}\{h(x_{i})\} \: and \: f = \gamma \circ \textbf{u}. \: Then,$

\vspace{6pt}

$(a) \: \forall S,\:  \mathcal{N}_{S}\subseteq \chi, \: f(T) = f(S) \: , if \: \mathcal{C}_{S}\subseteq T \subseteq \mathcal{N}_{S}$

\vspace{6pt}

$(b) \: \left | \mathcal{C}_{S} \right | \leq K$

\vspace{6pt}

$\forall S \in \chi$, $f(S)$ is determined by $\textbf{u}(S)$. For the $j_{th}$ dimension as the output of $\textbf{u}$, there exists at least one $x_{j} \in \chi$ such that $h_{j}(x_{j}) = \textbf{u}_{j}$, where $h_{j}$ is the $j$th dimension of the output vector from $h$. $\mathcal{C}_{S}$ is the union of all $x_{j}$ for $j = 1, ..., K$. Any additional points $x$ such that $h(x) \leq \textbf{u}(S)$ forms the union of $\mathcal{N}_{S}$.

This theorem implies that the neural network learns to recognize an object or a shape $S'$ by a set of critical points $\mathcal{C}_{S}$ from original $S'$. As long as this set of critical point $\mathcal{C}_{S}$ is maintained, the neural network can recognize this point cloud in general.

\subsection{Theoretical Analysis}

The core of our 3D multiview representation approach is to perform downsampling on the original 360$^\circ$ point clouds from various perspectives and viewpoints. We want to theoretically prove two sub-problems for this work: the effectiveness of our method in recognizing incomplete partial point clouds and improving the general performance on complete point clouds.

For the first sub-problem, it is quite probable that adding partial point sets will improve recognition effectiveness on partial sets, and it will be verified experimentally in section \ref{subsec:partialsceneseg}.   

We present the theoretical analysis for this second sub-problem below and suggest a simple but effective way to guarantee the improvement, which is to provide these multiview-based partial sets $\{S^{''}_{i}\}_{(i = 1,...,w)}$ as supplementary forms to the original complete training data $S^{'}$.

Based on Theorem 1, we suppose that for an object, a shape or a scene given, the ideal complete point set $S$ related to it should contain an infinite number of points. The current PointNet (and other DNNs for point clouds) deals with a still complete but relatively sparse point cloud $S^{'}$ which is a uniformly downsampled subset of $S$. Assume that $\mathcal{C}_{S}$ and $\mathcal{C}^{'}_{S}$ are respectively the critical point sets of $S$ and $S^{'}$. We further suppose that the value of K is fixed and is large enough, and we have $|\mathcal{C}^{'}_{S}| \leq |\mathcal{C}_{S}| \leq K$. 

We want to prove that by providing a multiview representation of complete point clouds to supplement the training data, i.e., the multiview-based partial point sets $\{S^{''}_{i}\}_{(i = 1,...,w)}$, to the current complete point sets $S^{'}$, we can in the worst case maintain the original performance on 360$^\circ$ complete point clouds or improve them by either maintaining the same $\mathcal{C}^{'}_{S}$ or bringing it closer to the ideal critical point sets $\mathcal{C}_{S}$. There are two different situations to discuss:

\textbf{(a) $\{S^{''}_{i}\}_{(i = 1,...,w)}$ is a multiview representation for $S^{'}$.}


$w$ is the number of partial point sets in this multiview presentation form for $S^{'}$. In this case, the points the network learn for 3D recognition actually remain the same, we have $S^{''}_{i} \subseteq S^{'}$, and therefore $\underset{i}{\cup }\{S^{''}\} \subseteq S^{'}$. Note that, $\forall x^{''} \in S^{''}_{i}$, $h_{j}(x^{''}) \leq \textbf{u}(S^{'})$, which implies that the critical point set $\mathcal{C}_{S}^{''}$ related to the union of partial point sets plus the original $S^{'}$ remains the same, equally for $\mathcal{N}^{''}_{S}$. Consequently, we have

\begin{equation}
\label{eq2}
    \mathcal{C}^{''}_{S} \equiv \mathcal{C}^{'}_{S} \: , \: \mathcal{N}^{''}_{S} \equiv \mathcal{N}^{'}_{S} 
\end{equation}

\textbf{(b) $\{S^{''}_{i}\}_{(i = 1,...,w)}$ is a multiview representation for $S$.}


$w$ is still the number of partial point sets in this multiview presentation form for $S$ and we have $S^{''}_{i} \subseteq S$.

If $\underset{i}{\cup }\{S^{''}\} \equiv S^{'}$, we have the same situation with the previous $(a)$, thus the $\mathcal{C}_{S}^{''}$ and $\mathcal{N}^{''}_{S}$ remain unchanged. Otherwise, if $\underset{i}{\cup }\{S^{''}\} \not\equiv S^{'}$, it is possible to have some changes in the $\mathcal{C}_{S}^{''}$ compared to $\mathcal{C}_{S}^{'}$. However, due to the property of $\mathcal{C}_{S}$ and max pooling layer, i.e. the $MAX$ operator,
for $\forall x_{j} \in \mathcal{C}_{S}$, $\forall x^{'}_{j} \in \mathcal{C}^{'}_{S}$ and $\forall x^{''}_{j} \in \mathcal{C}^{''}_{S}$, we have,

\begin{equation}
\label{eq3}
    h_{j}(x^{'}_{j}) \leq h_{j}(x_{j})
\: , \:
    h_{j}(x^{''}_{j}) \leq h_{j}(x_{j})
\end{equation}

Therefore, critical points change only when $h_{j}(x^{'}_{j}) < h_{j}(x^{''}_{j})$. Once the critical points are changed in this case, it actually brings the critical point set closer to the ideal critical point set $\mathcal{C}_{S}$. As a result, we have

\begin{equation}
\label{eq4}
    \begin{aligned}
    h_{j}(x^{'}_{j}) \leq h_{j}(x^{''}_{j}) \leq h_{j}(x_{j}) \\
    |\mathcal{C}^{'}_{S}| \leq  |\mathcal{C}^{''}_{S}| \leq |\mathcal{C}_{S}| \leq K \: , \: |\mathcal{N}^{'}_{S}| \leq  |\mathcal{N}^{''}_{S}| 
    \end{aligned}
\end{equation}

The above theoretical analysis implies the fact that we can at worst maintain the same overall performance on complete point clouds as without multiview representation, which is the case described in (a). However, PointNet's architecture determines the fact that it subsamples the input point clouds randomly to meet the preset requirement for point numbers in each point set in training,  we are thus most often in the second situation where the performance will usually be improved and it brings the effect of upsampling through multiview-based partial point sets which are actually obtained via downsampling. The analysis stands for general 3D recognition tasks on point clouds as well.

\section{3D Multiview Representation}
\label{sec:3drepresentation}

In this section, we present the concrete approach to realize 3D multiview representation operation for the existing 360$^\circ$ point clouds. We come back to the specific task of scene understanding where we seek to represent the complete scene point clouds as a collection of partial point sets. The overall framework is shown as Algorithm \ref{3d_multi}. 

\makeatletter
\def\BState{\State\hskip-\ALG@thistlm}
\makeatother

\begin{algorithm}[htb]
\caption{3D Multiview Representation for Scene Point Clouds}\label{3d_multi}
\begin{algorithmic}[1]
\State \textit{Define the Field of View(FOV) range}
\Procedure{Perspective Choice}{}
\State $\textit{Input scene data S'}$
\State $\textit{Grids distribution} \gets \textit{FOV}$
\State $\textit{Possible viewpoint V} \gets \textit{Intersections of gridlines}$
\State $\textit{Perspective} \gets \textit{V + angles of views} \: \theta$
\EndProcedure
\Procedure{Visibility}{}
\State $\textit{Visible points} \gets \textit{Perspective}$ 
\State $S''_{i} \gets \textit{Hidden Point Removal\cite{katz2007direct}}$
\EndProcedure
\State $\textit{3D Multiview representation} \gets \{S''_{i}\}_{(i = 1,...,w)}$ 
\end{algorithmic}
\end{algorithm}

\begin{figure}[htb]
\begin{center}
\includegraphics[width=0.95\linewidth]{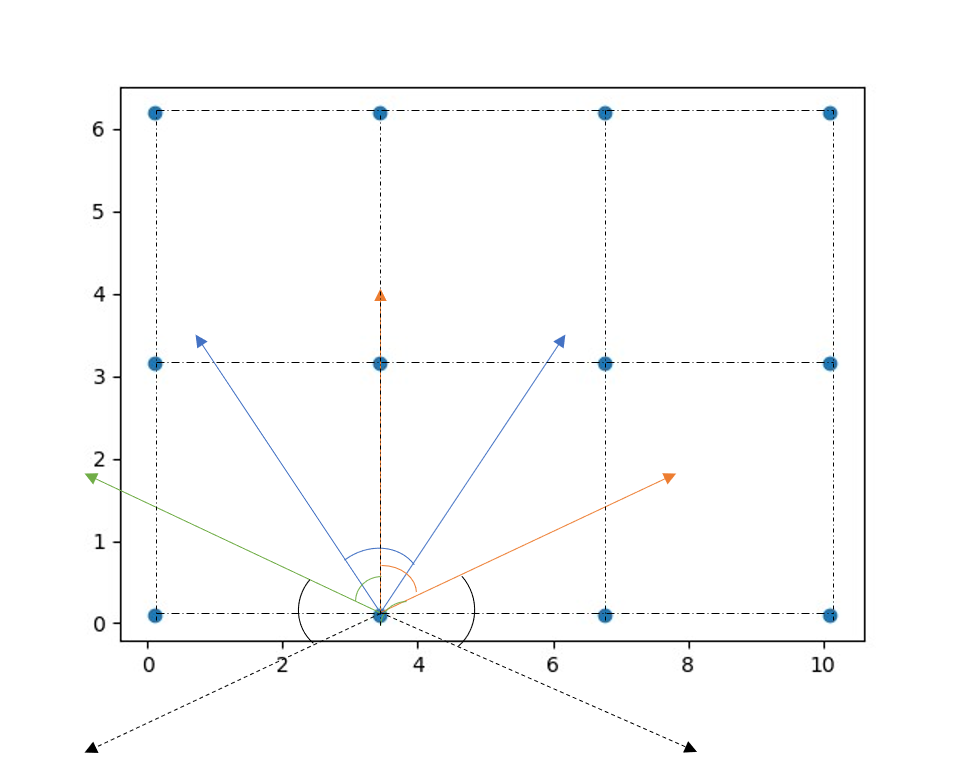}
\end{center}
    \caption{Illustration of the Perspective choice. The blue dots are the possible viewpoints, and the horizontal view range is presented by radial with different colors. Best viewed in color.}
    \label{fig:grid}
\end{figure}

We seek to simulate the multiview-based partial sets of point clouds as the actual partial point sets we will get with real cameras. The whole process can be separated into two stages, i.e. \textit{Perspective Choice} and \textit{Visibility Determination}. We firstly define the FOV. The parameters used come from Kinect, whose horizontal and vertical view ranges are respectively 70$^\circ$ and 60$^\circ$, and the valid depth ranges from 0.5 to 4m. We then split the entire scene data $S'$ into certain grids in the way that all area can be covered by the predefined FOV. The intersections of grid lines are regarded as possible viewpoints as in Fig. \ref{fig:grid}. Various angle of views, i.e. yaw and pitch, are assigned for the purpose of covering the whole scene.

When the Perspective$(V + \theta)$ is chosen, we downsample the complete scene data $S'$ to only visible points from the given Perspective choice. Hidden Point Removal(HPR)\cite{katz2007direct} is adopted to realize the visibility determination operation. The whole collection of $\{S''_{i}\}_{(i = 1,...,w)}$ is obtained by repeating the above process for each possible viewpoints and angle of views. An example of this multiview presentation is shown in Fig. \ref{fig:downsampled_example}. 

It is worth noting that we set a threshold for the minimum number of points contained in each multiview-based downsampled set of point clouds to eliminate some extremely sparse partial point set that contain fewer points than the threshold value. The threshold value is empirically set to 40,000. 


\section{Experiments and Discussions}
\label{sec:experiments}

\begin{figure*}[th]
\centering
\subfigure[Ground Truth label.]{
\begin{minipage}[t]{0.3\linewidth}
\centering
\includegraphics[width=2in,height=1.6in]{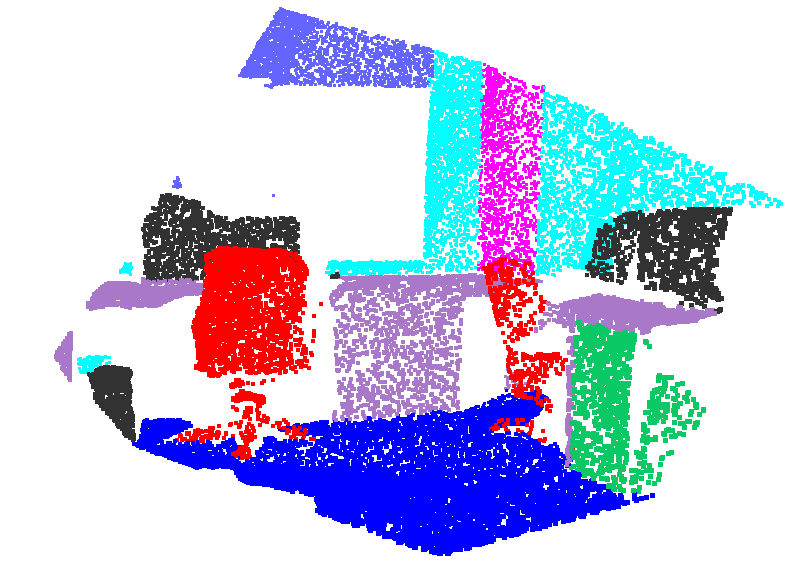}
\end{minipage}}
\subfigure[Baseline segmentation result.]{
\begin{minipage}[t]{0.3\linewidth}
\centering
\includegraphics[width=2in,height=1.6in]{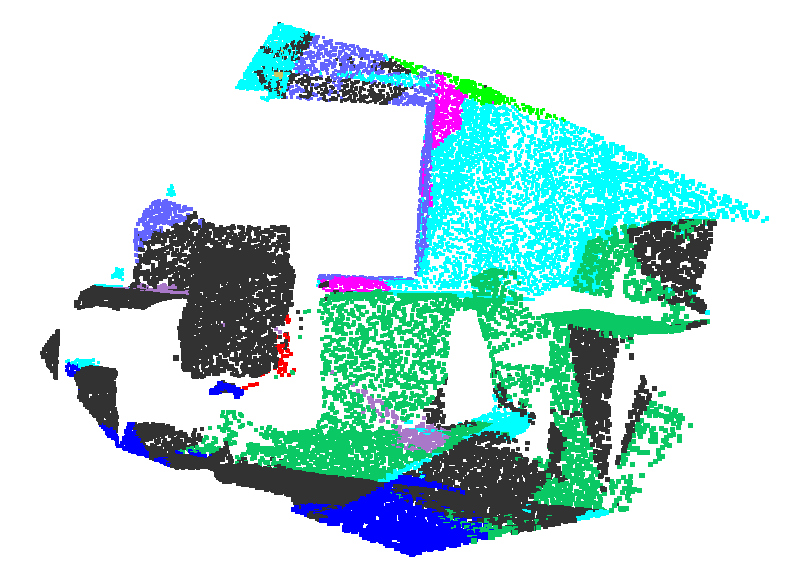}
\end{minipage}}
\subfigure[Segmentation result in EXP. 1.]{
\begin{minipage}[t]{0.3\linewidth}
\centering
\includegraphics[width=2in,height=1.6in]{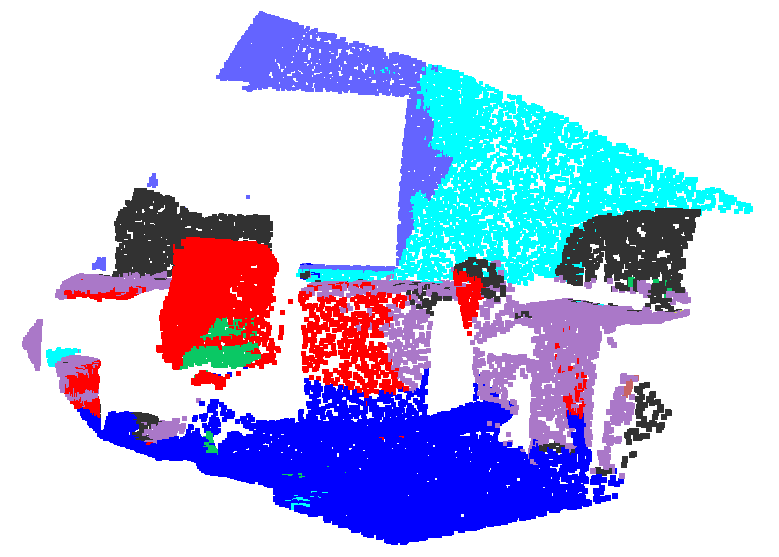}
\end{minipage}}
\caption{Ground truth label and segmentation results on a partial point set. Chairs in red, tables in purple, sofa in orange, board in gray, bookcase in green, floors in blue, windows in violet, beam in yellow, column in magenta, doors in khaki and clutters in black. Best viewed in color.}
\label{fig:seg_results_scene}
\end{figure*}

Experiments are divided into several parts. We firstly present the experimental results on partial scene semantic understanding task. Then we test our multiview representation method on 3D object part segmentation task. Finally, we analyze the trade-off between these multiview-based partial point clouds and original point clouds in the learning process of network.

\subsection{Partial Scenes Semantic Segmentation}
\label{subsec:partialsceneseg}

We use the Stanford dataset which covers an area of more than $6,000  m^{2}$ and contains over 215 million points\cite{armeni_cvpr16} in experiments. It has 11 distinct indoor scenes and 13 object categories, including both movable and immovable objects like sofa, chair, wall, floor etc., unspecified objects are classified as clutter. The whole dataset is divided into 6 areas including 272 rooms, the annotation is point-wise, each point has a label indicating its category. We follow the pre-processing steps in \cite{qi2017pointnet}, where each point is represented by a 9-dim vector of XYZ, RGB and normalized location as to the room (from 0 to 1).

The 3D multiview representation of each scene point set is obtained with the approach in section \ref{sec:3drepresentation}. We provide an overview of the original Stanford dataset and the corresponding 3D multiview representation in Table \ref{tab:stanford_overview}. For example, the Stanford dataset initially has 44 360$^\circ$ scene point clouds in Area 1, each of them shows the entire layout of a room, and those 44 scene point clouds are then downsampled and represented as 1164 multiview-based partial point sets.

\begin{table}[htb]
    \begin{center}
    \caption{Overview of the Stanford data and their multiview representations. The unit is the number of point sets.}
    \label{tab:stanford_overview}
    \scalebox{1.0}{
    \begin{tabular}{cccccccc}
    \hline
    \multicolumn{1}{c|}{Usage} & \multicolumn{5}{c|}{Training}   & \multicolumn{1}{c|}{Testing} &       \\ \cline{1-7}
\hline
Area        & 1    & 2    & 3   & 4    & 5    & 6       & Total \\
\hline
Original    & 44   & 40   & 23  & 49   & 68   & 48      & 272   \\
MV & 1164 & 1753 & 521 & 1165 & 2065 & 1270    & 7938  \\
\hline
    \end{tabular}}
    \end{center}
\end{table}

Area 6 is chosen as the testset while others are used for training. In order to provide a better comparison between the performance of the network before and after the fusion of 3D multiview representation data, we prepare two different testsets. The first testset contains the original 48 scene point clouds from Area 6, while the second one contains only the partial scene point sets, randomly chosen from the multiview representation data of Area 6. For each 360$^\circ$ scene point cloud in Area 6, we select two multiview-based partial point sets for testing, meaning that the second testset has in total 96 sets partial point clouds. 

We conduct several experiments. The first experiment setting is standard where the original complete point clouds from Area 1-5 are used for training, which is used as our baseline. In EXP. 1, we provide 100 incomplete partial sets from the multiview representation data for each area as supplement for training. Additionally, in order to guarantee the possible influence is not due to the increase in the total amount of training data but rather by the effect of these partial point sets, we include an extra experiment setting in which the training data size is reduced to maintain similar size of training data as in baseline experiment setting. This experiment is referred as EXP. 2. Basically, we remove all the point sets from Area 4 in the training process, including both complete and partial point clouds.



In Table \ref{tab:result semseg}, we compare the overall semantic segmentation results for each experimental setting on two different testsets, the number in bracket is the standard deviation for the accuracy calculated on each complete/partial scene. For the baseline (without the fusion of multiview representation), the performance on the testset 1 and testset 2 differs evidently. Here, the network performs poorly in recognizing the partial scenes. After supplementing the multiview-based partial point sets in EXP. 1, the performance on both testsets improve, indicating that multiview-based partial data help in improving the semantic segmentation results on both complete and partial point clouds. EXP. 2 further proves that this improvement is not due to the increase of the total amount of training data. More detailed segmentation results for each object category on both testsets are presented in Table \ref{tab:result class testset1} and Table \ref{tab:result class testset2}. The class-wise results are in accordance with the overall general results in Table \ref{tab:result semseg}. It is worth noting that the 3D multiview representation helps especially in improving the segmentation results on some movable objects like table, chair, sofa and bookcase. An example of segmentation results are given in Fig. \ref{fig:seg_results_scene}.

\begin{table}[htb]
\caption{Results on semantic segmentation. Metric is average classification accuracy(\%) for all the point clouds calculated on points.}
\label{tab:result semseg}
\begin{center}
\scalebox{1.0}{
\begin{tabular}{cccc}
\hline
Experiments & Baseline & EXP. 1 & EXP. 2 \\
\hline
Testset 1(complete)   & 86.1(0.059)   & \textbf{90.4}(0.039)   & 90.2(0.039)   \\
Testset 2(incomplete)   & 46.9(0.186)   & \textbf{78.8}(0.122)   & 77.1(0.101)  \\
\hline
\end{tabular}}
\end{center}
\end{table}

\begin{table*}[htb]
\caption{Results of testset 1 for each object class on semantic segmentation in scene. Metric is average IoU over each class.}
\label{tab:result class testset1}
\begin{center}
\scalebox{1.0}{
\begin{tabular}{c|c|ccccccccccccc}
\hline
Class  & mean          & ceiling       & floor         & wall          & beam          & column        & window        & door          & table         & chair         & sofa          & bookcase      & board         & clutter       \\
\hline
Baseline & 0.65          & \textbf{0.93} & 0.97          & 0.73          & 0.64          & 0.35          & 0.75          & 0.79          & 0.67          & 0.61          & 0.29          & 0.56          & 0.53          & 0.55          \\
EXP. 1 & \textbf{0.78} & \textbf{0.93} & \textbf{0.98} & \textbf{0.81} & \textbf{0.79} & \textbf{0.65} & 0.78          & 0.80          & \textbf{0.73} & \textbf{0.77} & \textbf{0.84} & 0.68          & \textbf{0.69} & \textbf{0.66} \\
EXP. 2 & 0.77          & \textbf{0.93} & 0.97          & 0.80          & \textbf{0.79} & 0.64          & \textbf{0.80} & \textbf{0.81} & \textbf{0.73} & 0.76 & 0.74          & \textbf{0.69} & 0.68          & 0.65         \\
\hline
\end{tabular}}
\end{center}
\end{table*}

\begin{table*}[t]
\caption{Results of testset 2 for each object class on semantic segmentation in scene. Metric is average IoU over each class.}
\label{tab:result class testset2}
\begin{center}
\scalebox{1.0}{
\begin{tabular}{c|c|ccccccccccccc}
\hline
Class  & mean          & ceiling       & floor         & wall          & beam          & column        & window        & door          & table         & chair         & sofa          & bookcase      & board         & clutter       \\
\hline
Baseline & 0.20          & 0.44          & 0.38          & 0.39          & 0.09          & 0.02          & 0.10          & 0.29          & 0.26          & 0.15          & 0.07          & 0.17          & 0.07          & 0.19          \\
EXP. 1 & \textbf{0.55} & \textbf{0.86} & \textbf{0.80} & \textbf{0.69} & \textbf{0.48} & \textbf{0.35} & \textbf{0.67} & \textbf{0.55} & \textbf{0.63} & \textbf{0.48} & 0.31          & 0.42          & \textbf{0.40} & \textbf{0.44} \\
EXP. 2 & 0.52          & 0.84          & 0.78          & 0.66          & 0.42          & 0.20          & 0.55          & \textbf{0.55} & \textbf{0.63} & \textbf{0.48} & \textbf{0.40} & \textbf{0.44} & 0.35          & 0.41    \\
\hline
\end{tabular}}
\end{center}
\end{table*}

\begin{table*}[t]
\begin{center}
\caption{Segmentation results on ShapeNet part dataset. areo means airplane. Metric is mIoU(\%) on points.}
\label{tab:shapenet}
\scalebox{0.87}{
\begin{tabular}{c|c|cccccccccccccccc}
\hline
                                                          & mean          & areo          & bag           & cap           & car           & chair         & \begin{tabular}[c]{@{}c@{}}ear\\ phone\end{tabular} & guitar        & knife         & lamp          & laptop        & motor         & mug           & pistol        & rocket        & \begin{tabular}[c]{@{}c@{}}skate\\ board\end{tabular} & table         \\
\hline
3DCNN                                                      & 79.4          & 75.1          & 72.8          & 73.3          & 70.0          & 87.2          & 63.5                                                & 88.4          & 79.6          & 74.4          & 93.9          & 58.7          & 91.8          & 76.4          & 51.2          & 65.3                                                  & 77.1          \\
PointNet                                          & 83.4          & 82.7          & 76.6          & 83.2          & 74.0          & 89.4          & 69.7                                                & 90.9          & 85.2          & 79.8          & 95.2          & 65.0          & 91.6          & 81.8          & 53.3          & 71.5                                                  & \textbf{80.9} \\
\hline
\begin{tabular}[c]{@{}c@{}}PointNet+\\our MV\end{tabular} & \textbf{83.9} & \textbf{83.3} & \textbf{79.5} & \textbf{86.2} & \textbf{76.8} & \textbf{90.0} & \textbf{74.2}                                       & \textbf{91.3}          & \textbf{85.7} & \textbf{80.2}          & \textbf{95.4}          & \textbf{66.8}          & \textbf{92.4} & \textbf{81.9}          & \textbf{60.5} & \textbf{73.7}                                         & 80.7 \\        
\hline
\end{tabular}}
\end{center}
\end{table*}

\subsection{3D Object Part Segmentation}

\begin{figure}[htb]
\begin{center}
    \includegraphics[width=2.8in]{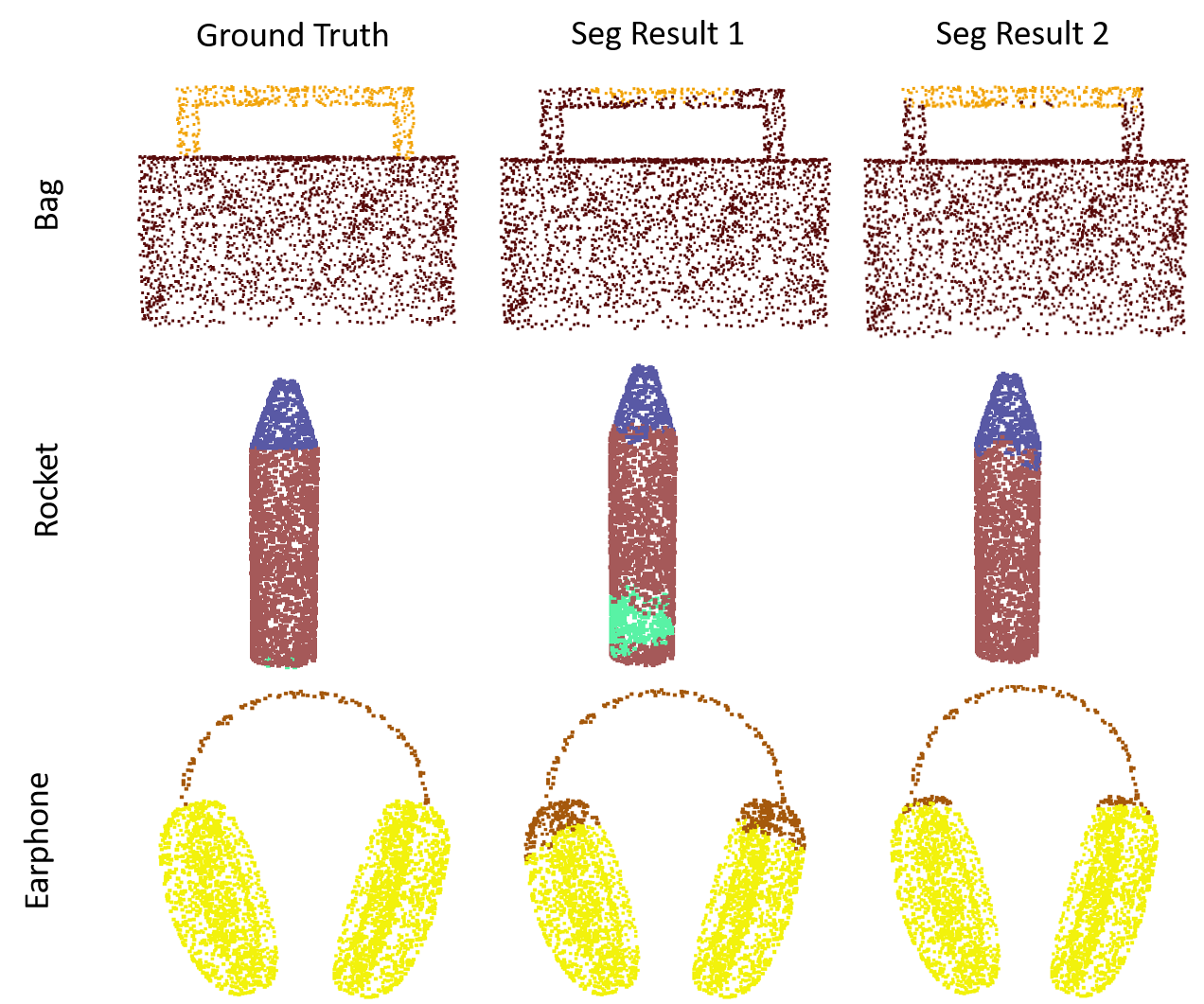}
\end{center}
    \caption{\textbf{Segmentation results on some of the object categories from ShapeNet.} The first column are the ground truth label, the second column are the part segmentation results from PointNet, while the last column shows the results after applying our 3D multiview representation. Best viewed in color.}
    \label{fig:shapent}
\end{figure}

As another challenging fine-grained 3D recognition task, part segmentation seeks to predict a part category label for each point on a 3D model. We evaluate our multiview-based processing method on the ShapeNet part dataset from \cite{chang2015shapenet}.The mIoU(\%)\footnote{mIoU: mean Intersection over Union.} on points is used as evaluation metric. We observe a performance improvement for 15 object categories out of 16, compared to the original PointNet. Some examples of segmentation results is given in Fig. \ref{fig:shapent}. Each category is evaluated in Table \ref{tab:shapenet}.\footnote{Note that the results reported here for PointNet is slightly different from \cite{qi2017pointnet}, because the authors published a slightly different version of source code on GitHub compared with the original code used in their paper. In order to make the comparison more objective, we reported the results obtained by the current version of code.}

\subsection{Trade-Off Relation}

\begin{figure}[htb]
    \begin{center}
    \includegraphics[width=3.2in]{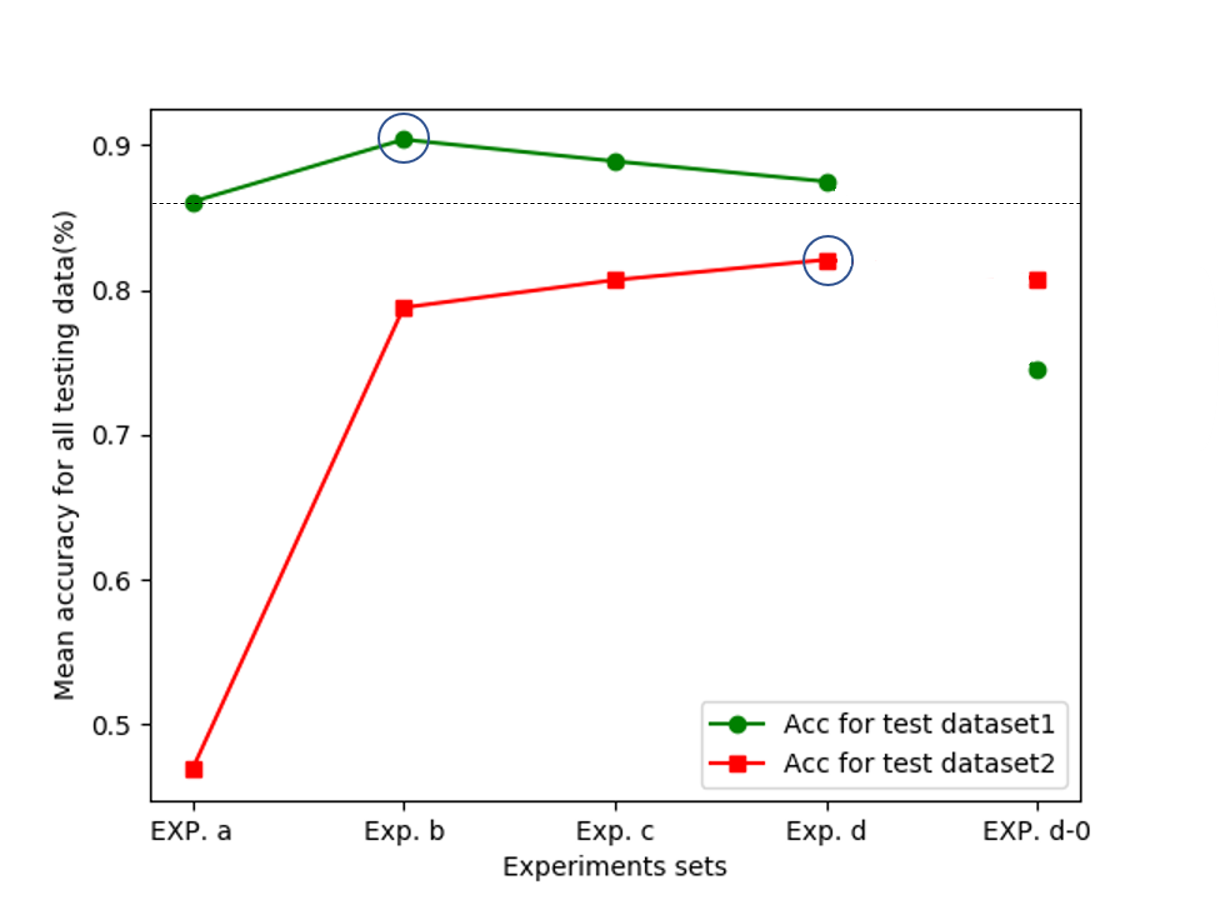}
    \end{center}
    \caption{Results on scene semantic segmentation with different amount of partial scene data added into training process.}
    \label{fig:tradeoff}
\end{figure}

While applying this 3D multiview representation method, we observe a trade-off between the number of these multiview-based partial point sets and their ability to recognize complete/incomplete partial point sets. In the beginning, with the increase in the size of partial point clouds supplemented in the training process, there is an improvement in the DNN's ability to recognize both complete and incomplete partial point clouds. However, if we continue to increase the size of the partial point data, the performance in segmenting complete point clouds will reach its peak and then begin to decrease while the performance on partial point sets continue to improve. As shown in Fig.\ref{fig:tradeoff}, we gradually increase the number of these partial point clouds into the training data, starting from 0, to 100, 200 and 300 partial sets of point clouds for each area, and they are respectively referred as EXP. a, EXP. b, EXP. c and EXP. d. We also design an extra experiment setting referred as EXP. d-0 where the original 360$^\circ$ are completely removed from the training date and replaced by the multiview-based partial scenes. We choose 300 partial point sets for each area in EXP. d-0, which means that the experiment setting is the same as EXP. d but without 360$^\circ$ point clouds.

It is worth noting that we always have a better performance on testset 1 than baseline when the partial scenes are supplemented.
And for EXP. d-0 that contains no complete point clouds, it is still able to reach comparable segmentation results for testset 2. According to the theoretical analysis, as long as the critical point set $\mathcal{C}^{'}_{S}$ for the 360$^\circ$ point cloud data $S'$ can be reconstructed by the partial point sets $S''_{i}$ from the 3D multiview representation, it is possible to completely replace the original training data set by the 3D multiview-based partial point sets and maintain the same performance. 


\section{Conclusion}
\label{sec:conclusion}
In this work, we propose a 3D multiview representation for point clouds in deep learning which helps to effectively improve the performance for scene semantic understanding on both 360$^\circ$ and partial point sets. This approach can be generalized for multiple other 3D recognition tasks like 3D parts segmentation.

Future work could look at incorporating this preprocessong approach into the architecture of the DNNs, which should make the whole process more robust and elegant.

\ifCLASSOPTIONcaptionsoff
  \newpage
\fi

\bibliographystyle{IEEEtran}
\bibliography{refs}             

\end{document}